\documentclass{article}
\usepackage[utf8]{inputenc}
\usepackage{hyperref}
\usepackage{geometry}
\usepackage{float}
\usepackage{graphicx}
\usepackage{multicol}
\usepackage{amssymb}
\usepackage{amsthm}

\usepackage{color}

\title{\textbf{A More Abstractive Summarization Model}}
\author{Satyaki Chakraborty (schakra1@cs.cmu.edu) \\ Xinya Li (xinyal@cs.cmu.edu) \\ Sayak Chakraborty (sayak.cs19@gmail.com)}
\date{May 2019}

\begin{document}

\maketitle

\begin{multicols}{2}

\section{Introduction}
Pointer generator networks \cite{see2017get} is an extremely popular method of text summarization. More recent works in this domain, like \cite{gehrmann2018bottom}, \cite{kryscinski2018improving} still build on top of the baseline pointer generator by augmenting a content selection phase, or by decomposing the decoder into a contextual network and a language model. In our work, we first thoroughly investigate why the pointer-generator network is unable to generate novel words, and then show that adding an Out-of-vocabulary (OOV) penalty, is able to improve the amount of novelty/abstraction significantly. We use normalized n-gram novelty scores from \cite{kryscinski2018improving} as a metric for determining the level of abstraction. Moreover, we also report rouge scores of our model since most summarization models are evaluated with R-1, R-2, R-L scores. 

\section{Related Work}
Research on text summarization can largely be divided into two approaches by methodology: supervised learning trained with a cross-entropy loss, and RL-based training which directly tries to optimize the ROUGE score. 
\subsection{Sequence-to-seqeuence models}
One of the pioneering work in the first approach is a seq2seq model \cite{nallapati2016abstractive} based on the
original encoder-decoder attention RNN model by \cite{bahdanau2014neural}. Its encoder is a
bidirectional GRU-RNN to form the condensed representation and the the decoder is a
unidirectional GRU-RNN which generates the summary. \cite{nallapati2016abstractive} proposes an original way to reduce the dimensionality of the softmax's output by thresholding the size of the vocabulary. \cite{nallapati2016abstractive} also applies a switching pointer-generator network to generate rare or out-of-vocab (OOV)
words during test time. \cite{nallapati2017summarunner} proposes a model for extractive summarization that uniquely models extractive summarization as sequence classification and learns
to generate extractive summaries even when the ground truths are abstract summaries.
The \textit{pointer-generator} model our project is based on by \cite{see2017get} shares the idea of generating OOV words by the network as in \cite{nallapati2016abstractive}. \cite{see2017get} design their network such that they would output \textit{hybrid} summaries of abstractive and extractive type and in the same time, avoid phrase repetition by incorporating a ‘coverage vector’ into the attention distribution and defining an auxiliary coverage loss that penalizes phrase repetition.

\subsection{RL based methods}
The main papers on applying RL methods to text summarization include \cite{paulus2017deep}, \cite{narayan2018ranking} and \cite{chen2018fast}. \cite{paulus2017deep} incorporates the standard encoder-decoder neural network with supervised
word prediction and reinforcement learning of policy using self-critical gradient. \cite{narayan2018ranking} poses an extractive summarization problem as a sentence ranking task and uses Reinforcement Learning for training the network by optimizing the ROUGE metric. \cite{narayan2018ranking} used an objective function that combines cross entropy loss with rewards obtained from policy gradient learning to optimize the ROUGE objective. \cite{chen2018fast} proposes a hybrid summarization task that extracts sentences and rewrite them abstractively while using hybrid network architecture and policy based reinforcement
learning.

\subsection{Methods to improve abstraction}
While most of the models focus on improving the ROUGE score of the generated summaries, \cite{kryscinski2018improving} is one of the few which attempt to actually focus on improving the level of abstraction. They do so by i) having a separate contextual network (which encodes the current state of the decoder) and a separate language model, and ii) adding a novelty reward optimized through policy gradient to encourage novel word generation. On the contrary, in our approach, we do not use any external optimization to reward novel word generation. Instead, we argue that this is a more fundamental issue and that novelty can be improved without an external reward.

\section{Datasets}
The most commonly used dataset for text summarization is the CNN / Daily Mail dataset by \cite{hermann2015teaching}. We used its processed version by \cite{nallapati2016abstractive} in our project, which also has become popular and was used to train \cite{see2017get}'s original \textit{pointer-generator} network. The original CNN / Daily Mail dataset contains online news articles (781 tokens on average) paired with multi-sentence summaries (3.75 sentences or 56 tokens on average)\cite{nlp}. The processed version contains 287,226 training pairs, 13,368 validation pairs and 11,490 test pairs\cite{nlp}. Both the original PG network and our model are evaluated with full-length F1-scores of ROUGE-1, ROUGE-2, ROUGE-L, and n-gram normalized novelty. 

Besides the CNN/DM dataset, we also use the Gigaword dataset(\cite{linguisticdataconsortium}) for survey material: it contains multiple summaries for the same source text and thus provides ample examples for our survey on people's preferences over different types, as we discuss below. 

\section{Survey}

The very motivation of our research project is to provide a better abstractive summarization tool. Although a vanilla pointer generator model aims to learn \textbf{hybrid} summaries between the abstractive and extractive type, its summaries still largely cling to the source text and contain relatively few novel words and a low novelty score or ``abstractiveness". To demonstrate its importance, we conducted a survey to investigate people’s preferences between abstractive and extractive ground-truth summaries from the Document Understanding Conference 2003 and 2004 summarization dataset\cite{duc2003}\cite{duc2004}. Each survey taker is presented with 5 pieces source texts and their abstractive and extractive summaries without being told which one is which. They are asked to choose the summary they prefer. A total of 15 takers completed the survey (75 preferences in total). Averaging the preferences over all choices with equal weight, 66.7\% are in favor of the abstractive summaries.

\section{Approach}
In this section, we firstly briefly discuss the details of the original vanilla pointer generator network to understand the fundamentals, and then shift to its limitations with our diagnosis. Finally, we show how to overcome those limitations with an additional loss and results of the modified algorithm. 
\end{multicols}
\begin{figure}[H]
    \center
    \includegraphics[width=\textwidth]{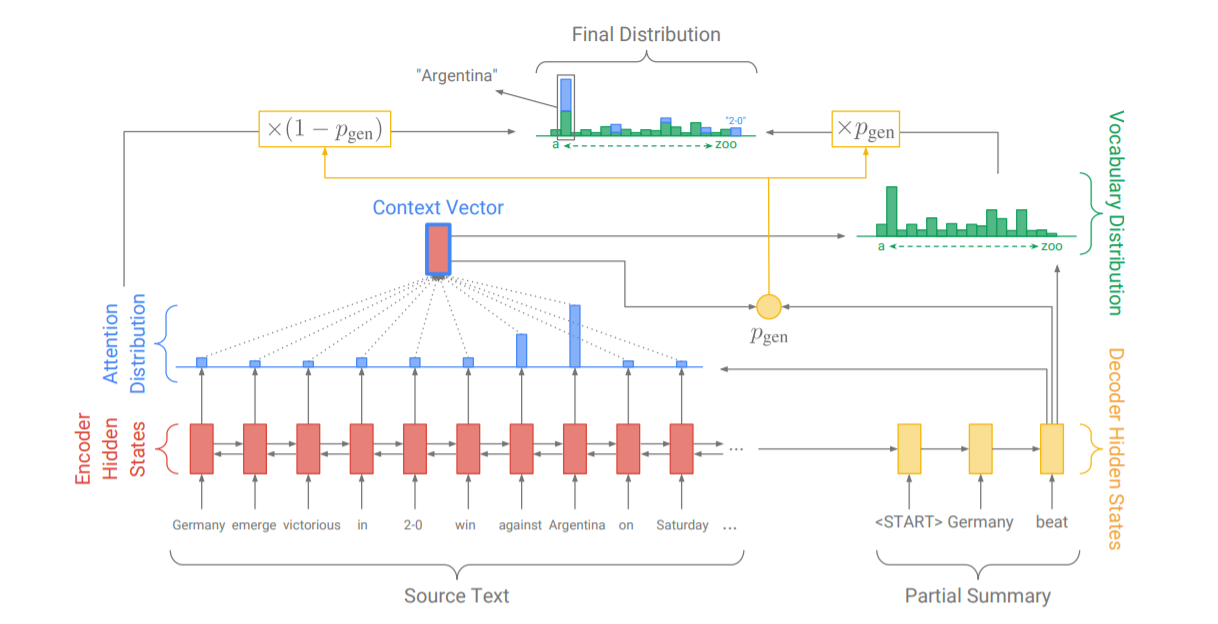}
    \caption{The network architecture of the vanilla \textit{pointer-generator}}
    \label{fig:my_label}
\end{figure}

\begin{multicols}{2}

\subsection{Vanilla model}
Our model is based on the \textit{pointer generator} network. In the model, a single-layer bidirectional LSTM layer encodes the source text and produces a sequence of encoder hidden states $h_i$. At every time step $t$, the unidirectional decoder receives $h_i$ and the current decoder state $s_t$ to generate the attention distribution as the probability of distribution over the \textit{source words}, $a^t$ (it makes the generation of OOV words inside source text possible) and a vocabulary distribution $p_{vocab}$
    \begin{equation}
        e^t_i = v^T tanh(W_hh_i + W_ss_t + b_{attn})
    \end{equation}
    \begin{equation}
        a^t = softmax(e^t)
    \end{equation}
where $v$, $W_h$, $W_s$ and $b_{attn}$ are learnable parameters. This attention distribution $p_{attn}$ given by $a^t$ is then subsequently used to generate a context vector $h_t^*$. The probability distribution over the entire vocabulary, or $p_{vocab}$ is also given as follows.
    \begin{equation}
    p_{vocab} = softmax(V^{'} (V[s_t, h_t^*] + b) + b)
    \end{equation}
, where $h_t^*$ is a context vector defined as:
    \begin{equation}
        h_t^* = \sum_{i} a_t^i h_i
    \end{equation}
The final probability distribution of word generation is a weighted combination between the attention distribution and the vocabulary distribution, where the respective ``out-of-domain" words have zero values (words not in source text and OOV words, respectively). Such combination weights are learned end-to-end and as \textit{generation probability}, or $p_{gen}$,
    \begin{equation}
        p_{gen} = \sigma(w^{T}_{h^{*}} h_{t}^* + w_s^Ts^t + w_x^T x^t + b_{ptr})
    \end{equation}
where vectors $w^{T}_{h^{*}}$, $w_s$, $w_x$ and scalar $b_{ptr}$ are learnable parameters and $\sigma$ is the sigmoid function. 
    \begin{equation}
        \label{eqn:comb}
        p_{final} = p_{gen} * p_{vocab} + (1-p_{gen}) * p_{attn}
    \end{equation}

We can easily see from Eq.(5) that $p_{gen}$ naturally acts like a control switch that decide whether the model will generate a new word from the \textit{vocabulary distribution} or \textit{source word distribution}. By controlling the generation probability of out-of-source-text words, $p_{gen}$ thus control the amount of abstraction of its summaries.

\subsection{Observations}
\subsubsection{Novelty}
We first calculate the unigram and bigram novelties of summaries generated by the vanilla pointer generator and the reference summaries to evaluate their quality. The novelty score is defined by \cite{kryscinski2018improving} as below: 
    \begin{equation}
        N(x^{gen}, n) = \frac{||ng(x^{gen}, n) - ng(x^{src}, n)||}{||ng(x^{gen}, n)||}
    \end{equation}
where $ng(x^{gen}, n)$ denotes the the function that computes the set of unique
n-grams in a document $x$; $x_{gen}$ denotes the generated summary; $x_{src}$ denotes the source document, and $||s||$ the number of words in a piece of text $s$ \cite{kryscinski2018improving}. A novel word is a word that is not in the given source text. Since abstractive summaries differ by the extractive ones exactly by its use of novel words, a novelty score effectively reflects on how abstractive a summary is compared to the source text. The novelty comparison table is shown below.

\begin{table}[H]
\begin{center}
\caption{Novelty Score Comparison between PG and Ground Truth}
\begin{tabular}{c||c|c}
      n-gram  & Pointer Generator & Ground Truth \\
     \hline 
    Unigram   & 0.07 & 2.23  \\
    \hline
    Bigram  & 13.5 & 49.97
\end{tabular}
\end{center}
\end{table}

As we can see from the table, the vanilla pointer generator network hardly produces any novel words. We next explore why this happens.

\subsubsection{$p_{gen}$ distribution}
The $p_{gen}$ distribution over 1000 randomly selected generated summaries is shown as follows. 
\begin{figure}[H]
    \centering
    \includegraphics[width=0.4\textwidth]{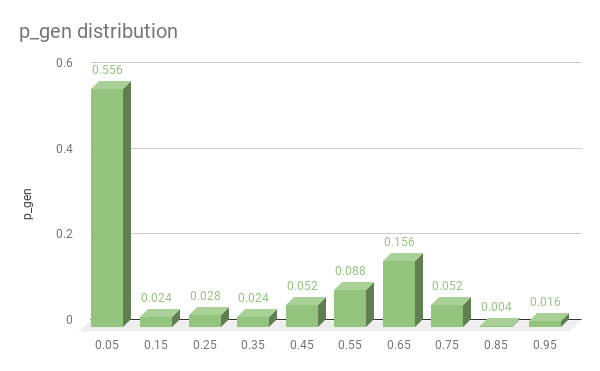}
    \caption{$p_{gen}$ distribution}
    \label{fig:pgen}
\end{figure}
From figure \ref{fig:pgen}, we observe that for more than 50 \% of the time, $p_{gen}$ is less than 0.1, and for less than 0.1 \% of the time, it is more than 0.7. This explains why novelty of the vanilla model is so less. Since $p_{gen}$ is extremely low most of the time, the final probability distribution is significantly biased towards the attention distribution and thus the model ends up mostly copying words from the source text and the vocabulary distribution is mostly just ignored. But why does $p_{gen}$ show this trend? To answer that question, we do the following analysis.

\subsubsection{$p_{vocab}$ vs $p_{attn}$ of sampled words}
We plot the $p_{vocab}$ and the $p_{attn}$ of every word that is sampled from the final distribution in a randomly selected summary. The scatter plot is shown as follows.
\begin{figure}[H]
    \centering
    \includegraphics[width=0.4\textwidth]{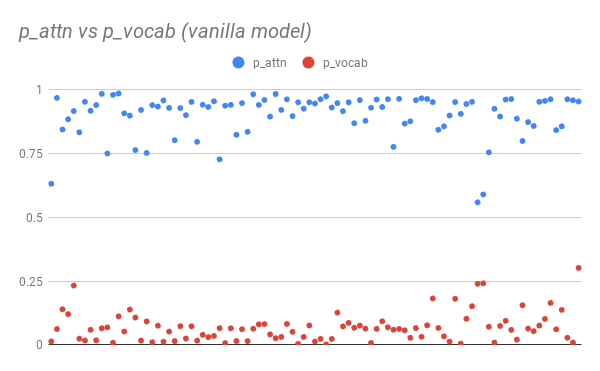}
    \caption{$p_{vocab}$ vs $p_{attn}$ of vanilla model. Time increases from left to right}
    \label{fig:vs0}
\end{figure}
We observe that even when a word exists in vocabulary, the contribution of the vocab distribution is non-significant. $p_{vocab}$ is hardly greater than 0.5 for all the words that have occurrence in the 50,000 word vocabulary. Since $ 0 \leq p_{vocab} \leq 1$, $ 0 \leq p_{attn} \leq 1$ and $p_{attn} > p_{vocab}$, from equation \ref{eqn:comb}, a higher value of $p_{final}$ is obtained when $p_{gen}$ is low. This explains the pattern shown in figure \ref{fig:pgen}. 

\textbf{But why does the network have such biases for the attention distribution?} We observe that in the initial phase of training, when the network is randomly initialized, both $p_{attn}$ and $p_{vocab}$ are low. As training proceeds, $p_{attn}$ slowly increases, but there is no significant change in $p_{vocab}$. This could be caused by the fact that $p_{attn}$ is distributed over 400 words (our maximum encoder sequence length), whereas $p_{vocab}$ is distributed over 50,000 words. It might be easier for the network to distribute the probability mass over 400 words compared to 50,000 words, and once the preference for $p_{attn}$ is set, $p_{gen}$ will start to increase to favour $p_{attn}$ over $p_{vocab}$ and thus the vocabulary distribution is always ignored.

\subsection{Modification}
Our assumption stems from a simple observation: named entities such as foreign people and place names often appear in the source text and are mostly out-of-vocabulary(OOV) words. The main reason why we have a separate copy distribution in the pointer generator network is because we would want to have the option of generating such OOV words in the summary by simply copying them from the source text. But for a non-OOV word, we argue that if we prefer the vocabulary distribution over the attention distribution, we would be able to increase the capability of the network to generate more novel words and thus result in more abstract summaries. This means, in our case, when a target word to be generated is OOV, we want $p_{gen}$ to be low (i.e. favour the attention distribution) and when a target word to be generated is non-OOV, we want $p_{gen}$ to be high. We formulate this constraint as an auxiliary loss which computes the negative log likelihood between $(1-p_{gen})$ and $y_{oov}$, where $y_{oov}=1$, if the target word is OOV and 0 otherwise.

    \begin{equation}
    \label{eqn:loss}
        L_{OOV} = -y_{oov}log(1-p_{gen}) - (1-y_{oov})log(p_{gen})
    \end{equation}

\begin{figure}[H]
    \centering
    \includegraphics[width=0.4\textwidth]{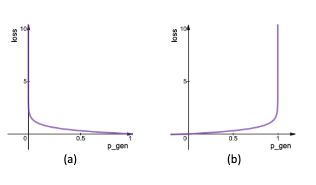}
    \caption{$L_{oov}$. (a) $L_{oov}$ when target word is out-of-vocabulary, a lower value of $p_{gen}$ is preferred. (b) $L_{oov}$ when target word is not OOV, a higher value of $p_{gen}$ is preferred.}
    \label{fig:loss}
\end{figure}

\textbf{It is to be noted that,} even though we have penalized the model to have a preference for vocabulary distribution, for a non-OOV word, whether we should use the vocab distribution or the copy distribution is debatable. In order to prevent the model from solely relying on the vocabulary distribution for non-OOV words, we train the network in three phases, first vanilla pointer generator without coverage loss, second with coverage loss and third with the auxiliary loss in \ref{eqn:loss}. This gives us a reasonable balance between the 2 distributions as is observed in the following figure.

\begin{figure}[H]
    \centering
    \includegraphics[width=0.4\textwidth]{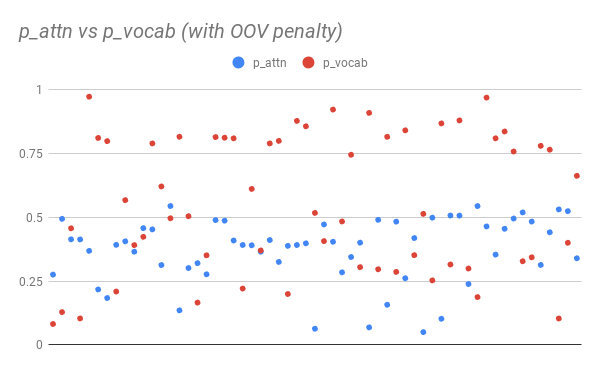}
    \caption{$p_{vocab}$ vs $p_{attn}$ of our model. Time increases from left to right.}
    \label{fig:vs1}
\end{figure}

\section*{Results}
Before we jump into our result report, it is necessary to note that there has not yet been a robust metric to evaluate abstractive text summaries. Metrics like ROUGE have serious limitations\cite{nlp}: i) metrics like ROUGE only assess content selection and do not account for other quality aspects, such as factual accuracy and fluency; ii) these metrics rely mostly on lexical overlap to evaluate content selection, which naturally puts an abstractive summary in \textit{disadvantage}, as what makes abstractive summary different is that it summarizes the source text without perfect lexical overlaps. To address this problem, we choose to use normalized n-gram novelty defined in by Eq.(7)\cite{kryscinski2018improving}. 

Having decided on an evaluation metric effective for abstractive summaries, given the low $p_{gen}$ distribution of the original pointer-generator model, we hypothesized that our modified model with penalty would score much higher in novelty by comparison. To assess our hypothesis, we plot the novelty score comparison between the vanilla pointer-generator model, our model and the ground truth in order below. Each single novelty score is calculated by normalizing over 1000 randomly-chosen samples from each source. The plot matches our hypothesis that our summaries show a huge improvement on novelty score over the original model. In both cases of unigram and bigram, our score is more than midway between the original one and the ground truth's, which unsurprisingly scores the highest. Our much higher novelty scores, together with the mixed $p_{vocab}$ and $p_{attn}$, demonstrate an increase of abstractiveness in our summaries.

\begin{figure}[H]
    \centering
    \includegraphics[width=0.4\textwidth]{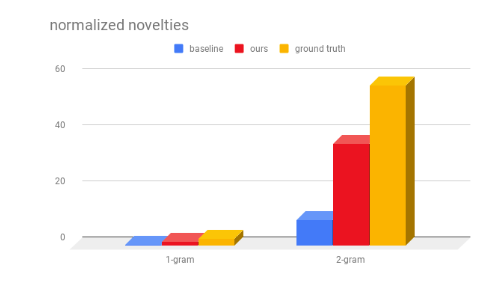}
    \caption{Normalized novelties of baseline vs ours vs ground-truth}
    \label{fig:novelty}
\end{figure}

We also report our ROUGE scores with the original model's below, as ROUGE score is included by almost all other summarization tools as one of their evaluation metric -- albeit it is not ideal for ours. 
\begin{table}[H]
\begin{center}
\caption{Rouge Score Comparison between PG and Our Modified PG}
\begin{tabular}{c||c|c}
        & Pointer Generator & Our Modified PG \\
     \hline 
    R-1   & 38.82 & 35.67  \\
    \hline
    R-2  & 15.74 & 12.9 \\
    \hline 
    R-l  & 35.91 & 33.28
\end{tabular}
\end{center}
\end{table}

\section{Future Directions}
Our results of mixed $p_{attn}$ and $p_{vocab}$ demonstrates a definitive improvement of summaries' abstractiveness. It is worth mentioning that there is trade-off between ROUGE score and novelty. As the novelty increases, the ROUGE score should decrease. For this report, due to time and resource constraints, we were unable to measure the correlation between the two, and analyze the number of fine-tuning iterations needed to hit the sweet spot. For better understanding of the trade-off, we plan to plot of the pareto frontier between n-gram novelty and ROUGE-n for different fine tuning iterations.

\bibliographystyle{IEEEtran}


\end{multicols}

\end{document}